\documentclass[11pt,a4paper]{article}
\usepackage[hyperref]{emnlp-ijcnlp-2019}
\usepackage{times}
\usepackage{latexsym}

\usepackage{url}
\usepackage{graphicx}
\usepackage{amsmath}
\usepackage{amssymb}
\usepackage{booktabs}
\usepackage{CJKutf8}
\usepackage{multirow}
\usepackage{lipsum}
\usepackage[marginal]{footmisc}

\aclfinalcopy % Uncomment this line for the final submission

\title{Lexical Sememe Prediction using Dictionary Definitions by Capturing Local Semantic Correspondence}

\author{Jiaju Du, Fanchao Qi, Maosong Sun, Zhiyuan Liu\\
Department of Computer Science and Technology, Tsinghua University\\
Institute for Artificial Intelligence, Tsinghua University\\
Beijing National Research Center for Information Science and Technology\\
{\tt \{djj18,qfc17\}@mails.tsinghua.edu.cn,\{sms,lzy\}@tsinghua.edu.cn} \\}

\date{}
\newcommand\blfootnote[1]{
    \begingroup
    \renewcommand\thefootnote{}\footnote{#1}
    \addtocounter{footnote}{-1}
    \endgroup
}

\begin{document}
\maketitle
\begin{abstract}
Sememes, defined as the minimum semantic units of human languages in linguistics, have been proven useful in many NLP tasks. 
Since manual construction and update of sememe knowledge bases (KBs) are costly, the task of automatic sememe prediction has been proposed to assist sememe annotation. 
In this paper, we explore the approach of applying dictionary definitions to predicting sememes for unannotated words. 
We find that sememes of each word are usually semantically matched to different words in its dictionary definition, and we name this matching relationship \textit{local semantic correspondence}.
Accordingly, we propose a Sememe Correspondence Pooling (SCorP) model, which is able to capture this kind of matching to predict sememes. 
We evaluate our model and baseline methods on a famous sememe KB HowNet and find that our model achieves state-of-the-art performance. 
Moreover, further quantitative analysis shows that our model can properly learn the local semantic correspondence between sememes and words in dictionary definitions, which explains the effectiveness of our model. 
The source codes of this paper can be obtained from \url{https://github.com/thunlp/scorp}.\blfootnote{\noindent This paper is accepted by \textit{Journal of Chinese Information Processing} (in Chinese). This version is for English readers.}
\end{abstract}

\section{Introduction}
In linguistics, \textit{sememes} are defined as the minimum semantic units of human languages \cite{bloomfield1926set}.
Some linguists believe that the meanings of all the words can be described with a limited closed set of sememes, which is similar to the idea of semantic prime \cite{wierzbicka1996semantics}.
However, sememes are normally implicit and cannot be recognized directly. Therefore, people manually annotate words with a set of predefined sememes to construct sememe KBs. 

\begin{figure}
    \centering
    \includegraphics[width = 0.48\textwidth]{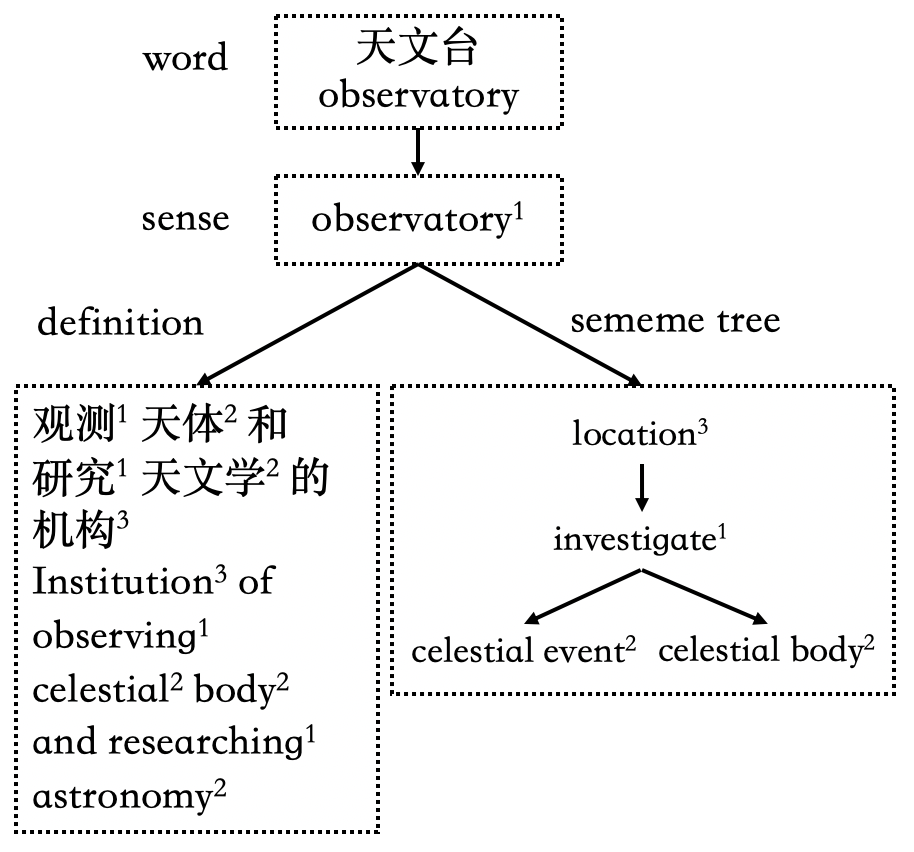}
    \caption{The annotation of word ``observatory'' in HowNet and its definition in dictionary. Sememes and definition words with the same superscript are semantically matched.}
    \label{fig:hownet-example}
\end{figure}

HowNet \cite{dong2003HowNet} is the most famous sememe KB. 
It defines about $2,000$ language-independent sememes and uses them to annotate over 100 thousand Chinese and English words. 
Every word in HowNet contains some senses, and each sense is annotated with a hierarchical structure of sememes, i.e., sememe tree. 
Figure \ref{fig:hownet-example} exhibits an example of how words are annotated in HowNet.
The word ``observatory'' has one sense, and this sense has 4 sememes, namely ``location'', ``investigate'', ``celestial body'' and ``celestial event''.
HowNet has been successfully utilized in various NLP applications such as semantic similarity computation \cite{liu2002}, word sense disambiguation \cite{zhang2005chinese, duan2007word}, sentiment analysis \cite{zhu2006semantic, xianghua2013multi, dang2010method}, language modeling \cite{gu2018language}, word embedding \cite{niu2017improved}, and word classification \cite{zeng2018chinese}. 

Seeing that numerous new words emerge constantly and meanings of existing words keep altering, it is labor-intensive and time-consuming to expand and update sememe KBs like HowNet.
To solve this problem, \citet{xie2017lexical} propose the task of sememe prediction, aiming to automatically recommend related sememes to the words without sememe annotation.
They also present two simple but effective embedding-based models for the task.
These two models ignore the hierarchical structure of sememe trees and predict a set of sememes for every word.
\citet{jin2018incorporating} further propose to incorporate Chinese characters of words into sememe prediction and improve the prediction performance. 

These methods rely heavily on the representations of words or characters. 
As a result, it is hard to predict proper sememes for those low-frequency words or words with low-frequency characters because of their poor embeddings. 
In fact, there are other available resources that can be used in sememe prediction to tackle the challenge. 
Dictionary definitions clearly explain the meanings of words including low-frequency words, and they are also easy to obtain. 
Thus, we believe dictionary definitions are appropriate for sememe prediction.
\citet{li2018sememe} firstly propose to use dictionary definitions in sememe prediction.
However, they adopt a sequence to sequence model that foists inappropriate order on sememes. Additionally, they regard definitions as sequences of characters, which can hurt representations of the definitions owing to Chinese characters' ambiguity.

In this paper, we propose a novel model for sememe prediction using dictionary definitions. 
The model not only addresses the issues of existing methods but also can capture \textit{local semantic correspondence}, a kind of particular matching relationship between sememe trees and definitions.
More explicitly, for a given word, we find each of its sememes can be matched to one or more of its definition words, i.e., words in the dictionary definition.
Taking Figure \ref{fig:hownet-example} for example, the sememes ``location'', ``investigate'', and ``celestial body, celestial event'' are semantically matched to the words ``institution'', ``observing, researching'', and ``celestial, astronomy'' respectively. 

The reason for local semantic correspondence is easy to explain. 
Both a word's sememe tree and dictionary definition express the same meaning. 
The sememe tree of ``observatory'' represents the meaning of ``location of investigating celestial events and celestial bodies'', which is almost identical to that of the dictionary definition. 
Furthermore, sememe trees and dictionary definitions can be decomposed into sememes and definition words respectively. 
Consequently, it is natural to see that there is local semantic correspondence between a word's sememes and definition words.

To take advantage of local semantic correspondence, we propose the model \textbf{Sememe Correspondence Pooling (SCorP)}. 
SCorP computes the correspondence scores between all sememes and words in definitions, and do max-pooling over correspondence scores for every sememe.
Moreover, SCorP model ignores the hierarchical structures of sememes in sememe prediction following previous works \cite{xie2017lexical,jin2018incorporating,li2018sememe}, and uses the framework of sequence to set multi-label classification to avoid imposing order on sememes. 
And the inputs to SCorP model are word sequences rather than character sequences, which remedies the ambiguity problem of characters and at the same time, enables utilizing the sememe information of the definition words. 
In addition, we propose two effective retrofitting operations for SCorP which improves the prediction performance. 
In our experiments, we evaluate the sememe prediction performance of our SCorP model and several baselines. 
Experimental results show that our model achieves state-of-the-art performance. 
We also conduct further quantitative analysis and find that our model can correctly match definition words to sememes, which explains the superiority of our model.

To conclude, the contributions of this paper are twofold: (1) we discover \textit{local semantic correspondence}, a kind of specific semantic matching between a word's sememe tree and definition; (2) we propose a novel model which utilizes local semantic correspondence to predict sememes, and it achieves state-of-the-art performance in sememe prediction.

\section{Related Work}

\paragraph{Applications of Sememes} 
Sememes have been widely used in NLP tasks, such as semantic similarity computation \cite{liu2002}, word sense disambiguation \cite{zhang2005chinese, duan2007word}, and sentiment analysis \cite{xianghua2013multi, dang2010method, zhu2006semantic, huang2014new}. 
In addition, \citet{niu2017improved} incorporate sememes into word representation learning and utilize sememes to capture the exact meaning of a word within specific contexts.
\citet{zeng2018chinese} use sememe knowledge with the attention mechanism to determine the type of a word and expand the Chinese LIWC lexicon \citep{pennebakerlinguistic}. 
\citet{gu2018language} propose to regard sememes as linguistic experts to help predict suitable words in language modeling.

\paragraph{Sememe Prediction} \citet{xie2017lexical} present the automatic sememe prediction task for the first time. They also propose a collaborative filtering-based model SPWE and a matrix factorization-based model SPSE for this task and achieve acceptable sememe prediction results.
Following this work, \citet{jin2018incorporating} propose two models SPWCF and SPCSE which can leverage the internal character information of Chinese words, as well as an ensemble model CSP which considers both internal and external information. 
\citet{li2018sememe} first explore the role of dictionary descriptions in sememe prediction. 
They propose a sequence to sequence model named LD+Seq2Seq, where the inputs are character sequences of dictionary definitions or wiki descriptions, and the outputs are sequences of predicted sememes. 
In addition, there are also works trying to build sememe KB for another language by cross-lingual sememe prediction \cite{qi2018cross}. 

\paragraph{Applications of Dictionary Definitions} 
Dictionary definitions are abundant and valuable research resources in NLP. 
They have been used in various tasks, such as word sense disambiguation \cite{luo2018incorporating}, knowledge representation learning \cite{xie2016representation,zhong2015aligning}, reading comprehension \cite{long2017world}, and reverse dictionary \cite{hill2016learning,bosc2018auto}. 
Most of the previous works encode definitions into vectors for downstream tasks. 
In addition, some works adopt graph-based methods to build a word graph using dictionary definitions for downstream tasks \cite{thorat2016implementing,tissier2017dict2vec}.

\section{Methodology}
In this section, we first introduce some terms and notations which will be used below. Then we give a brief description to a basic sequence to set multi-label classification framework, on which our SCorP model is based. Next, we present our SCorP model and two different retrofitting operations in detail. Finally, we succinctly describe an ensemble model.

\subsection{Terms and Notations}
As mentioned previously, we use ``definition word'' to refer to a word in dictionary definitions. And we use ``target word'' to signify the word for which we want to predict sememes.
We define $W$ as the vocabulary set and $S$ as the set of sememes. 
Given a word $w\in W$, all of its sememes form a set $S_w=\{s_1,...,s_{|S_w|}\}\subseteq S$, where $|\cdot|$ represents the cardinality of a set. 
The dictionary definition of word $w$ is denoted by $D_w=\{d_1,...,d_{|D_w|}\}$, where $d_i$ is its $i$-th definition word. 
We use lowercase boldface symbols for vectors and uppercase boldface symbols for matrices. For example, $\mathbf{d}_i$ is the word vector of $d_i$, $\mathbf{s}_i$ is the sememe vector of $s_i$, and $\mathbf{D}_w$ is the matrix composed of word vectors $\mathbf{d}_1,..., \mathbf{d}_{|D_w|}$. 

\subsection{Basic Multi-label Classification Framework}
\label{section:basic}
The sequence to set multi-label classification (MC) framework serves as the base of our SCorP model.
It has two main parts, the encoder which can encode a dictionary definition into a vector, and the multi-label classifier, which uses the definition vector to compute association scores for each sememe. 
And the sememes with high scores are selected as the predicted sememes. 

We choose Bidirectional LSTM (BiLSTM) \cite{schuster1997bidirectional} as the encoder.
Formally, for the dictionary definition $D_w=\{d_1,...,d_{|D_w|}\}$ of a target word $w$, we pass $\{\mathbf{d}_1,...,\mathbf{d}_{|D_w|}\}$, the pre-trained word embeddings of definition words, to the BiLSTM. 
Then BiLSTM will output two sequences of hidden states:
\begin{equation}
    \begin{aligned}
        \{\stackrel{\rightarrow}{\mathbf{h}}_1,...,\stackrel{\rightarrow}{\mathbf{h}}_{|D_w|}\},\{\stackrel{\leftarrow}{\mathbf{h}}_1,...,\stackrel{\leftarrow}{\mathbf{h}}_{|D_w|}\}\\
        ={\rm BiLSTM}(\mathbf{d}_1,...,\mathbf{d}_{|D_w|}).
    \end{aligned}
    \label{equation:bilstm}
\end{equation}
We use the concatenation of the last hidden states of both directions as the definition vector which is denoted by $\mathbf{v}$, and feed it to a fully connected layer:
\begin{equation}
    \begin{aligned}
        \mathbf{v}&={\rm Concatenate}(\stackrel{\rightarrow}{\mathbf{h}}_{|D_w|},\stackrel{\leftarrow}{\mathbf{h}}_1), \\
        \mathbf{x}&=\mathbf{W}\mathbf{v}+\mathbf{b},
    \end{aligned}
    \label{equation:a}
\end{equation}
where $\mathbf{W}\in \mathbb{R}^{|S|\times 2l},\mathbf{x},\mathbf{b}\in \mathbb{R}^{|S|}$, $l$ represents the dimension of hidden states in a single direction. 
$[\mathbf{x}]_j$, the $j$-th element of $\mathbf{x}$, denotes the association score of $j$-th sememe. 
For training, we use the multi-label one-versus-all cross-entropy loss:
\begin{equation}
    L=-\frac{1}{|S|}\sum_{j=1}^{|S|}[\mathbf{y}_j]\sigma([\mathbf{x}]_j)+(1-[\mathbf{y}_j])\sigma(-[\mathbf{x}]_j),
\end{equation}
where $[\mathbf{y}]_j\in \{0,1\}$ represents whether the $j$-th sememe is in the sememe set of word $w$.

\subsection{SCorP Model}
\begin{figure}[t]
    \centering
    \includegraphics[width = 0.48\textwidth]{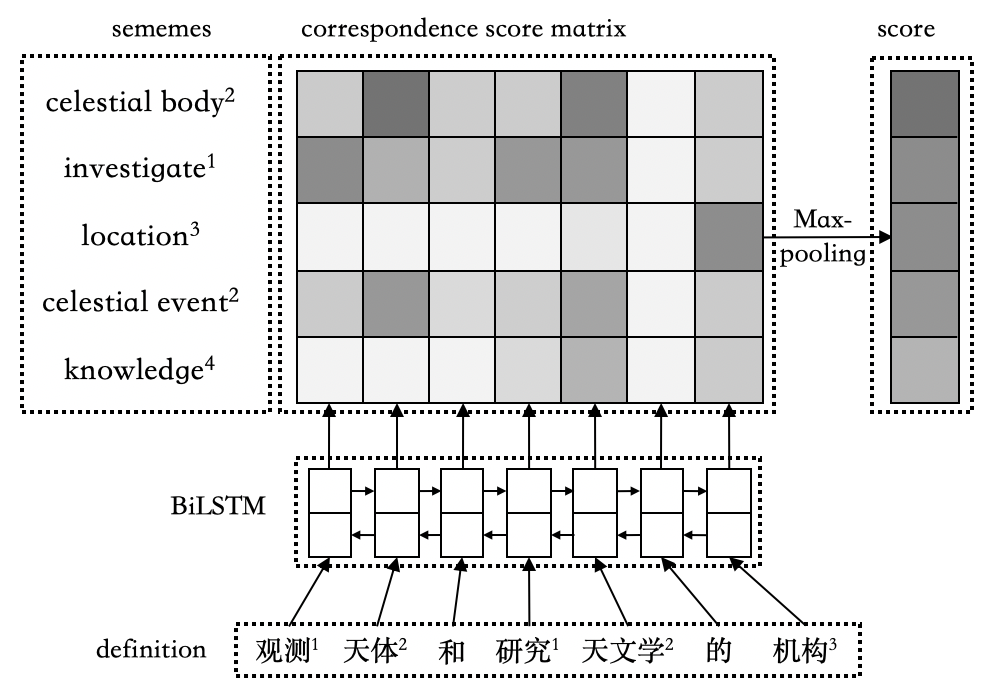}
    \caption{A sememe prediction example of SCorP model. The deeper a cell of the correspondence score matrix is, the higher the correspondence score between the sememe and the definition word. See Figure \ref{fig:hownet-example} for English translation of the definition. }
    \label{fig:sememe-correspondence-pooling}
\end{figure}
\label{section:scp}
The basic MC model encodes the whole definition into a vector, which will cause information loss and especially, make it difficult to cope with long definitions. 
Our Sememe Correspondence Pooling (SCorP) model, which exploits \textit{local semantic correspondence}, can handle this problem.

A sememe prediction example of SCorP model is illustrated in Figure \ref{fig:sememe-correspondence-pooling}. 
The encoder of SCorP model is similar to that of MC model. 
But SCorP model uses the concatenations of all the bidirectional hidden states in Equation \ref{equation:bilstm}, rather than only the concatenation of the last bidirectional hidden states:
\begin{equation}
    \begin{aligned}
        \mathbf{h}_i={\rm Concatenate}(\stackrel{\rightarrow}{\mathbf{h}}_i,\stackrel{\leftarrow}{\mathbf{h}}_i), 1\le i\le |D_w|.
    \end{aligned}
\end{equation}
Obviously $\mathbf{h}_i$ contains the context semantic information of $d_i$, the $i$-th definition word. 
Then every $\mathbf{h}_i$ is passed to a fully connected layer, and the output constitutes a matrix:
\begin{equation}
    \begin{aligned}
        \mathbf{y}_i&=\mathbf{W}\mathbf{h}_i+\mathbf{b}, \\
        \mathbf{Y}&=(\mathbf{y}_1,...,\mathbf{y}_{|D_w|}), \\
    \end{aligned}
\end{equation}
where $\mathbf{Y}\in \mathbb{R}^{|S|\times |D_w|}$ is in fact the semantic correspondence matrix, and $[\mathbf{Y}]_{ji}$ measures the correspondence between $j$-th sememe and $i$-th definition word.

According to \textit{local semantic correspondence}, each sememe of a word strongly corresponds to its one or more definition words. 
In other words, in sememe prediction, whether a sememe is selected as the predicted result depends on whether there is strong correspondence between the sememe and some words in the definition. 
Inspired by this, SCorP model calculates the final association score of every sememe by doing max-pooling over the correspondence scores between the sememe and all the definition words:
\begin{equation}
    \begin{aligned}
        \left[\mathbf{x}\right]_j=\max_{1\le i\le |D_w|}[\mathbf{y}_i]_j.
    \end{aligned}
\end{equation}

\subsection{Retrofitting Operations}
\label{section:retrofitting}
\subsubsection*{Incorporating Sememes of Definition Words}
In this subsection, we propose to incorporate sememe information of definition words into the encoder. 
On the one hand, sememe information has been proven effective in improving word embeddings \cite{niu2017improved}. On the other hand, we find that if a sememe corresponds to a definition word, it usually belongs to the sememe set of the definition word. 
Still taking the target word ``observatory'' for example, its sememe ``investigate'' semantically corresponds to the definition word ``observing'' and at the same time, ``investigate'' is also annotated to ``observing'' in HowNet.
Therefore, the sememes of definition words should be helpful to sememe prediction. 

We use a simple but effective method to incorporate sememe information of definition words, i.e., adding the average of sememe embeddings to the embedding of the definition word:
\begin{equation}
    \begin{aligned}
        \mathbf{d}'_i=\mathbf{d}_i+\frac{1}{|S_{d_i}|}\sum_{s_j\in S_{d_i}}\mathbf{s}_j,
    \end{aligned}
\end{equation}
where $S_{d_i}$ is the sememe set of $d_i$ and $\mathbf{d}'_i$ is the retrofitted embedding of $d_i$. 
For the definition words without sememe annotation, their retrofitted embeddings are equal to their original word embeddings. For simplicity, we use ``+SE'' to indicate incorporating sememes of definition words.

\subsubsection*{Integrating Embeddings of Target Words}
\label{section:target}
In this subsection, we present an approach to integrating embeddings of target words into our model. 
If a target word has word embedding, we simply put it and a separator in front of its dictionary definition to acquire the retrofitted definition sequence. 
We choose colon (:) as the separator. 
Formally, for a given target word $w$, the retrofitted definition sequence is $\{w, :, d_1,...,d_{|D_w|}\}$. 
Then we feed the embeddings of the elements of the retrofitted definition sequence to the BiLSTM encoder.

As for the target words that have no embedding, we borrow the idea of byte pair encoding from sequence modeling \cite{sennrich2016neural} to deal with them. 
For a target word without embeddings, we first split it into smaller subwords or characters which have embeddings. Then we add these subwords or characters and a separator to the definition sequence of the target word. We use ``+TW'' to indicate integrating embeddings of target words.

\subsection{Ensemble Model}
Our SCorP model utilizes dictionary definitions and word embeddings, while previous CSP model \cite{jin2018incorporating} mainly uses internal character information of target words. 
The two models use diverse information and naturally, we combine the two models to build an ensemble model. This model scores a sememe by doing a weighted addition of sememe scores outputted by the two submodels. 

\section{Experiments}
\subsection{Dataset}
\label{section:dataset}
We choose HowNet as the sememe KB, which annotates 103,843 Chinese words and 103,390 English words. 
These words contain 212,539 language-independent senses.
Following previous works \cite{xie2017lexical,jin2018incorporating}, we disregard the hierarchical structures of sememes in sememe prediction and filter out the low-frequency sememes which appears less than $5$ times in HowNet. 
The final number of sememes we use is $1,400$. 
In addition, we concatenate all definitions of a word and use the concatenated definition for training and evaluation.

We evaluate our models on Chinese and English. 
For Chinese, we use \textit{Modern Chinese Dictionary (6th Edition)}\footnote{\url{http://www.cp.com.cn/book/978-7-100-08467-3\_44.html}} as the source of dictionary definitions. 
It defines about $69,000$ Chinese words and characters. 
We extract dictionary definitions and segment them into words using THULAC \cite{li2009punctuation}. 
The final dataset contains $48,383$ words, which simultaneously have sememe annotation, word embeddings, and dictionary definitions. Word embeddings are trained using Skip-gram \cite{mikolov2013distributed} on the SogouT corpus\footnote{Sogou-T is a corpus of web pages provided by a Chinese commercial search engine. \url{https://www.sogou.com/labs/resource/t.php}}. 
The dataset is randomly divided into non-overlapping training, validation, and test sets in the ratio of 8:1:1. 

For English, we use WordNet \cite{miller1995wordnet} as the source of dictionary definitions, and use the pre-trained Glove \cite{pennington2014glove} word embeddings. The dataset has 43,907 words and is also divided in the ratio of 8:1:1. Table \ref{tab:statistics} shows some statistics of the two dataset.

\begin{table}
    \begin{center}
    \resizebox{\linewidth}{!}{
    \begin{tabular}{ccc}
        \toprule
         & Chinese & English \\
        \midrule
        \#words in dictionary & 69,000 & 147,306 \\
        \#senses in dictionary & 98,732 & 206,978 \\
        \#words in HowNet & 103,843 & 103,390 \\
        \#words with embeddings & 5,606,977 & 400,000 \\
        \#words in dataset & 48,383 & 43,907 \\
        avg. sememes of words & 2.57 & 3.00 \\
        \bottomrule
    \end{tabular}
    }
    \end{center}
    \caption{Some statistics of the two datasets}
    \label{tab:statistics}
\end{table}

\subsection{Experimental Setup}

\paragraph{Baseline Methods} 
We choose the following 4 models as baselines:
(1) SPWE \cite{xie2017lexical}, an embedding-based model. It first finds some similar words for the target word in the embedding space, then recommends sememes of similar words.
(2) CSP \cite{jin2018incorporating}, an ensemble model utilizing both internal character information, which is employed by its submodels SPWCF and SPCSE, and external word information, which are employed by its submodels SPWE and SPSE \cite{xie2017lexical};
(3) LD+Seq2Seq \cite{li2018sememe}, which utilizes dictionary definitions by a reformed sequence to sequence model; 
(4) MC, the basic sequence to set multi-label classification model in section \ref{section:basic}. 

\paragraph{Hyper-parameters and Training} 
For all the models, the dimension of word embeddings is $200$.
For SCorP and MC models, the dimension of BiLSTM hidden states is $256\times 2$. 
For all the other baseline methods, their hyper-parameters except word embedding dimension are respectively tuned to the best on the validation set. 
For our ensemble model, the weight ratio of correspondence scores is $\lambda_{\rm SCorP}:\lambda_{\rm CSP}=1:10$, which is also determined by tuning on the validation dataset. 
As for training, we use Adam Optimizer \cite{kingma2014adam} and the learning rate is $0.001$. During the training process, the word embeddings are frozen. Dropout is employed in SCorP and MC.

\paragraph{Evaluation Protocol} 
Following previous works, we use mean average precision (MAP) and F1 score as evaluation metrics. For a word with $K$ sememes, the model will score all sememes and rank them by scores. If the rankings of the $K$ sememes are $p_1\le p_2\le ...\le p_K$, we can compute MAP by
\begin{equation}
    MAP=\sum_{k=1}^K\frac{k}{p_k}
\end{equation}
For F1 score, we set a threshold $\delta$ for all models except LD+Seq2Seq. All sememes with scores higher than $\delta$ compose the output set. $\delta$ is a hyper-parameter and also tuned to the best on the validation set. We conduct 10-fold cross-validation in experiments. 

\subsection{Overall Sememe Prediction Performance}
\label{section:main-result}

\begin{table}[t!]
    \begin{center}
    \resizebox{\linewidth}{!}{
    \begin{tabular}{ccccc}
        \toprule
        \multirow{2}*{Models} & \multicolumn{2}{c}{Chinese} & \multicolumn{2}{c}{English} \\
         & MAP & F1 & MAP & F1 \\
        \midrule
        SPWE & 55.04 & 48.23 & 40.56 & 38.25 \\
        CSP & 58.93 & 50.26 & 42.58 & 37.56 \\
        LD+Seq2Seq & 30.49 & 33.28 & 24.63 & 28.70 \\
        \midrule
        MC & 51.24 & 42.06 & 49.09 & 39.71 \\
        MC(+TW) & 59.15 & 48.77 & 54.56 & 45.34 \\
        MC(+SE) & 53.99 & 45.90 & 52.62 & 45.24 \\
        MC(+TW,SE) & 60.55 & 50.84 & 56.57 & 48.95 \\
        \midrule
        SCorP & 54.95 & 46.89 & 56.17 & 50.41 \\
        SCorP(+TW) & 63.46 & 53.07 & 59.70 & 52.89 \\
        SCorP(+SE) & 57.57 & 49.99 & 58.28 & 52.83 \\
        SCorP(+TW,SE) & \textbf{64.65} & \textbf{54.62} & \textbf{61.53} & \textbf{55.22} \\
        \midrule
        Ensemble & \textbf{69.19} & \textbf{57.99} & \textbf{63.60} & \textbf{56.50} \\
        \bottomrule
    \end{tabular}
    }
    \end{center}
    \caption{Overall sememe prediction results of all the models on the test set. }
    \label{tab:main-results}
\end{table}

Table \ref{tab:main-results} lists the sememe prediction performance of our models and baseline methods. From this table, we can see that: 

(1) Our proposed SCorP(+TW,SE) model achieves the best sememe prediction performance compared with all the single models and previous ensemble model CSP. 
In addition, our proposed ensemble model achieves state-of-the-art performance.

(2) Our SCorP model and its variants always perform substantially better than the MC model and its corresponding variants.
This suggests that a size-fixed definition vector is not enough to contain all the information of the dictionary definition, and it is necessary to consider all the hidden states of the definition. 
Moreover, the max-pooling operation can effectively capture the local semantic correspondence. 
To validate this, we compare max-pooling with mean-pooling and attention mechanism in SCorP. 
All of the three operations can collect information from the hidden states of BiLSTM.
The mean-pooling averages the correspondence score for every sememe. 
In the model with attention mechanism, for every sememe, we use the sememe vector to attend over all the hidden states of the LSTM. The output of the attention is used to compute the association score of the sememe by a fully-connected layer.
After replacing the max-pooling with mean-pooling and attention mechanism, the MAP drops from 54.95 to 51.76 and 52.23 respectively.
The reason is that a sememe of the target word usually semantically corresponds to only a few words and has nothing to do with most of definition words. 
In the models with mean-pooling or attention mechanism, the association score of a sememe is inevitably affected by all the hidden states of the definition words, and the hidden states of irrelevant words bring lots of noise. 

(3) Basically, both +SE and +TW enhance the sememe prediction performance, no matter for MC or SCorP model. 
+TW gains most, because word embeddings are powerful and previous works relying only on target word embeddings have achieved acceptable results.

(4) LD+Seq2Seq model has relatively poor performance. This demonstrates it is not a good idea to impose inappropriate order on sememes. Furthermore, using character sequence as input is another possible reason for the bad performance, because the ambiguity of characters may hurt the precise representations of definitions.

\subsection{Sememe Prediction for OOV Words}
\begin{table}
    \begin{center}
    \begin{tabular}{ccc}
        \toprule
        Model & IV & OOV \\
        \midrule
        CSP & 59.03 & 42.92 \\
        LD+Seq2Seq & 29.33 & 28.15 \\
        \midrule
        SCorP & 54.44 & 51.05 \\
        SCorP(+TW) & \textbf{63.10} & \textbf{54.30} \\
        SCorP(+TW-WS) & 62.97 & 51.20 \\
        \bottomrule
    \end{tabular}
    \end{center}
    \caption{Sememe prediction MAPs for in-vocabulary (IV) and out-of-vocabulary (OOV) words of different models. +TW-WS signifies removing word splitting in the +TW operation.}
    \label{tab:phrase-results}
\end{table}
The models which heavily rely on target word embeddings don't work or suffer serious performance deterioration when faced with out-of-vocabulary (OOV) words. 
Our SCorP model can cope with the OOV words because it needs no target word embeddings. 
And the word splitting workaround of our +TW operation can also mitigate the OOV problem. 

To test the ability to handle the OOV problem of existing models, we expand the original dataset by adding $1,595$ OOV words which have no embeddings. Then we evaluate our models and baseline methods on the expanded dataset.
Notice that for CSP, only two submodels SPWCF and SPCSE can predict sememes for OOV words, and other two submodels are not used.

The evaluation results are exhibited in Table \ref{tab:phrase-results}. From the table, we can observe that our plain SCorP model and its variant SCorP(+TW) dramatically outperform the two baseline methods, especially for the OOV words. 
This manifests our model's extraordinary generalization capability to tackle the OOV problem. 
We also confirm the effectiveness of word splitting workaround by a lesion experiment. After removing word splitting for target words in +TW (denoted by +TW-WS), SCorP(+TW,-WS) model performs considerably worse in sememe prediction for the OOV words, even close to the plain SCorP model making no use of target word embeddings. 

\subsection{Influence of Word Frequency}
\begin{table}[t]
    \begin{center}
    \resizebox{\linewidth}{!}{
    \begin{tabular}{ccccc}
        \toprule
         Word Frequency & $\ge$5000 & 500-5000 & 50-500 & $\le$50 \\
        \midrule
        CSP & 62.18 & 59.60 & 46.60 & 26.34 \\
        LD+Seq2Seq & 29.31 & 31.83 & 33.61 & 31.93 \\
        \midrule
        MC(+TW,SE) & 63.18 & 61.75 & 59.08 & 54.32 \\
        SCorP(+TW,SE) & 65.42 & 64.71 & 62.94 & 58.91 \\
        \bottomrule
    \end{tabular}
    }
    \end{center}
    \caption{Sememe prediction MAPs for words with different frequency ranges. The numbers of words in the four ranges are 31481, 10714, 4528 and 1660 respectively. }
    \label{tab:frequency-results}
\end{table}
Table \ref{tab:frequency-results} lists the evaluation results of 4 models for words with different frequencies.
SCorP(+SE,TW) produces the best performance in all the word frequency ranges, which manifests its remarkable robustness. 
CSP, which relies heavily on word embeddings, suffers a sharp performance drop when faced with low-frequency words (word frequency less than 50), although they perform acceptably on high-frequency words.
LD+Seq2Seq is less affected by word frequencies, but it badly lags behind other models on whichever word frequency range.

\subsection{Case Study}
\label{section:case-study}
\begin{CJK*}{UTF8}{gbsn}
\begin{table*}[t!]
    \begin{center}
    \resizebox{\linewidth}{!}{
    \begin{tabular}{cl}
        \toprule
        Definition Words & Predicted Sememes \\
        \midrule
        天文台 (observatory) & celestial body/0.20, tell/-2.60, celestial event/-3.26, images/-4.63, light/-4.65, knowledge/-5.00 \\
        ： & celestial body/-4.37, tell/-5.66, time/-5.74, part/-6.38, morning/-7.08, past/-7.17 \\
        观测 (observe) & \textbf{investigate/2.37}, celestial event/-0.20, far/-1.28, celestial body/-1.35, measurement/-2.19, look/-2.68 \\
        天体 (celestial body) & \textbf{celestial body/6.16}, \textbf{celestial event/1.68}, investigate/0.63, measurement/-2.12, far/-3.40, look/-3.51 \\
        和 (and) & celestial body/-1.95, investigate/-2.66, celestial event/-3.61, find/-4.26, look/-5.32, choose/-5.86 \\
        研究 (research) & investigate/1.78, celestial event/-0.88, celestial body/-1.99, research/-2.27, find/-2.62, knowledge/-3.46 \\
        天文学 (astronomy) & celestial body/2.50, investigate/1.54, celestial event/1.29, knowledge/0.56, daytime/-1.31, earch/-1.85 \\
        的 (of) & part/-5.56, time/-6.31, human/-6.36, place/-7.04, tell/-7.16, head/-7.74 \\
        机构 (institution) & \textbf{location/1.97}, celestial event/-1.32, knowledge/-1.48, celestial body/-2.02, facility/-2.14, tool/-2.90 \\
        。 & time/-1.66, celestial body/-3.02, part/-3.21, investigate/-4.90, celestial event/-4.95, daytime/-5.46 \\
        \midrule
        Predicted Sememes & \textbf{celestial body/6.16}, \textbf{investigate/2.37}, \textbf{location/1.97}, \textbf{celestial event/1.68}, knowledge/0.56, far/-1.28 \\
        \bottomrule
    \end{tabular}
    }
    \end{center}
    \caption{Sememe prediction case for the word ``observatory''. Each row of the first column, except the last row, is filled with a definition word of ``observatory'', and the second column lists top three sememes and their correspondence scores for the corresponding definition word. The last row lists the final predicted sememes and their correspondence scores after max-pooling. The sememes in boldface means correct sememes. }
    \label{tab:case-study}
\end{table*}
\end{CJK*}
In this section, we conduct a case study on a sememe prediction example, to illustrate the effectiveness of our SCorP model in capturing local semantic correspondence. 

We still pick the word ``observatory'', which has four sememes involving ``location'', ``investigate'', ``celestial event'' and ``celestial body''. 
Our SCorP model obtains the correspondence matrix measuring the semantic correspondence between each sememe and each definition word. 
Table \ref{tab:case-study} shows the six most corresponding sememes and their correspondence scores for each definition word. 
If a sememe and its correspondence score in a certain row is in boldface, it denotes that the sememe achieves the highest correspondence score on the definition word in this row.
For example, on the row of definition word ``institution'', sememe ``location'' is in boldface, which means ``location'' corresponds with ``institution'' most closely. 
In addition, sememes ``investigate'' and ``celestial body, celestial event'' corresponds with words ``observe'' and ``celestial body'' most closely, respectively.
The semantic correspondence scores between all punctuation and all sememes are relatively low.
These results demonstrate that our SCorP model can properly capture the local semantic correspondence.
The last row lists the predicted sememes with the highest correspondence scores after max-pooling. 
We can clearly see that the four correct sememes obtain the highest scores and are ranked top 4 exactly.

\section{Conclusion and Future Work}
In this paper, we discover the particular \textit{local semantic correspondence} in sememe prediction using dictionary definitions. Accordingly, we propose a novel model, which can exploit this property to predict sememes. And we also design two useful retrofitting operations, including incorporating sememes of definition words and integrating embeddings of target words. Experimental results show that our model with the two retrofitting operations achieves state-of-the-art performance. 

We will explore the following research directions in the future: (1) taking into account the hierarchical structures of sememes and allocating context-dependent weights
when incorporating sememes of definition words.  
(2) predicting the hierarchical structure of sememes as well, which is a key step for expanding and updating sememe KBs like HowNet. Moreover, dictionary definitions are suitable for structured sememe prediction because different types of definition words correspond to the sememes in different hierarchical layers.
(3) transferring our model to cross-lingual sememe prediction to assist in the building of sememe KBs for other languages. 
\bibliography{emnlp-ijcnlp-2019}

\begin{thebibliography}{33}
\expandafter\ifx\csname natexlab\endcsname\relax\def\natexlab#1{#1}\fi

\bibitem[{Bloomfield(1926)}]{bloomfield1926set}
Leonard Bloomfield. 1926.
\newblock A set of postulates for the science of language.
\newblock \emph{Language}, 2(3):153--164.

\bibitem[{Bosc and Vincent(2018)}]{bosc2018auto}
Tom Bosc and Pascal Vincent. 2018.
\newblock Auto-encoding dictionary definitions into consistent word embeddings.
\newblock In \emph{Proceedings of EMNLP}.

\bibitem[{Dang and Zhang(2010)}]{dang2010method}
Lei Dang and Lei Zhang. 2010.
\newblock Method of discriminant for {Chinese} sentence sentiment orientation
  based on {HowNet}.
\newblock \emph{Application Research of Computers}, 4:43.

\bibitem[{Dong and Dong(2003)}]{dong2003HowNet}
Zhendong Dong and Qiang Dong. 2003.
\newblock {HowNet}-a hybrid language and knowledge resource.
\newblock In \emph{Proceedings of International Conference on Natural Language
  Processing and Knowledge Engineering}.

\bibitem[{Duan et~al.(2007)Duan, Zhao, and Xu}]{duan2007word}
Xiangyu Duan, Jun Zhao, and Bo~Xu. 2007.
\newblock Word sense disambiguation through sememe labeling.
\newblock In \emph{Proceedings of IJCAI}.

\bibitem[{Fu et~al.(2013)Fu, Liu, Guo, and Wang}]{xianghua2013multi}
Xianghua Fu, Guo Liu, Yanyan Guo, and Zhiqiang Wang. 2013.
\newblock Multi-aspect sentiment analysis for {Chinese} online social reviews
  based on topic modeling and {HowNet} lexicon.
\newblock \emph{Knowledge-Based Systems}, 37:186--195.

\bibitem[{Gu et~al.(2018)Gu, Yan, Zhu, Liu, Xie, Sun, Lin, and
  Lin}]{gu2018language}
Yihong Gu, Jun Yan, Hao Zhu, Zhiyuan Liu, Ruobing Xie, Maosong Sun, Fen Lin,
  and Leyu Lin. 2018.
\newblock Language modeling with sparse product of sememe experts.
\newblock In \emph{Proceedings of EMNLP}.

\bibitem[{Hill et~al.(2016)Hill, Cho, Korhonen, and Bengio}]{hill2016learning}
Felix Hill, Kyunghyun Cho, Anna Korhonen, and Yoshua Bengio. 2016.
\newblock Learning to understand phrases by embedding the dictionary.
\newblock \emph{Transactions of the Association for Computational Linguistics},
  4:17--30.

\bibitem[{Huang et~al.(2014)Huang, Ye, Wang, Chen, Cheng, and
  Zhu}]{huang2014new}
Minlie Huang, Borui Ye, Yichen Wang, Haiqiang Chen, Junjun Cheng, and Xiaoyan
  Zhu. 2014.
\newblock New word detection for sentiment analysis.
\newblock In \emph{Proceedings of ACL}.

\bibitem[{Jin et~al.(2018)Jin, Zhu, Liu, Xie, Sun, Lin, and
  Lin}]{jin2018incorporating}
Huiming Jin, Hao Zhu, Zhiyuan Liu, Ruobing Xie, Maosong Sun, Fen Lin, and Leyu
  Lin. 2018.
\newblock Incorporating {Chinese} characters of words for lexical sememe
  prediction.
\newblock In \emph{Proceedings of ACL}.

\bibitem[{Kingma and Ba(2014)}]{kingma2014adam}
Diederik~P Kingma and Jimmy Ba. 2014.
\newblock Adam: A method for stochastic optimization.
\newblock \emph{arXiv preprint arXiv:1412.6980}.

\bibitem[{Li et~al.(2018)Li, Ren, Dai, Wu, Wang, and Sun}]{li2018sememe}
Wei Li, Xuancheng Ren, Damai Dai, Yunfang Wu, Houfeng Wang, and Xu~Sun. 2018.
\newblock Sememe prediction: Learning semantic knowledge from unstructured
  textual wiki descriptions.
\newblock \emph{arXiv preprint arXiv:1808.05437}.

\bibitem[{Li and Sun(2009)}]{li2009punctuation}
Zhongguo Li and Maosong Sun. 2009.
\newblock Punctuation as implicit annotations for {Chinese} word segmentation.
\newblock \emph{Computational Linguistics}, 35(4):505--512.

\bibitem[{Liu and Li(2002)}]{liu2002}
Qun Liu and Sujian Li. 2002.
\newblock Word similarity computing based on {HowNet}.
\newblock \emph{International Journal of Computational Linguistics and Chinese
  Language Processing}, 7(2):59--76.

\bibitem[{Long et~al.(2017)Long, Bengio, Lowe, Cheung, and
  Precup}]{long2017world}
Teng Long, Emmanuel Bengio, Ryan Lowe, Jackie Chi~Kit Cheung, and Doina Precup.
  2017.
\newblock World knowledge for reading comprehension: Rare entity prediction
  with hierarchical lstms using external descriptions.
\newblock In \emph{Proceedings of EMNLP}.

\bibitem[{Luo et~al.(2018)Luo, Liu, Xia, Chang, and Sui}]{luo2018incorporating}
Fuli Luo, Tianyu Liu, Qiaolin Xia, Baobao Chang, and Zhifang Sui. 2018.
\newblock Incorporating glosses into neural word sense disambiguation.
\newblock In \emph{Proceedings of ACL}.

\bibitem[{Mikolov et~al.(2013)Mikolov, Sutskever, Chen, Corrado, and
  Dean}]{mikolov2013distributed}
Tomas Mikolov, Ilya Sutskever, Kai Chen, Greg~S Corrado, and Jeff Dean. 2013.
\newblock Distributed representations of words and phrases and their
  compositionality.
\newblock In \emph{Proceedings of NIPS}.

\bibitem[{Miller(1995)}]{miller1995wordnet}
George~A Miller. 1995.
\newblock Wordnet: a lexical database for english.
\newblock \emph{Communications of the ACM}.

\bibitem[{Niu et~al.(2017)Niu, Xie, Liu, and Sun}]{niu2017improved}
Yilin Niu, Ruobing Xie, Zhiyuan Liu, and Maosong Sun. 2017.
\newblock Improved word representation learning with sememes.
\newblock In \emph{Proceedings of ACL}.

\bibitem[{Pennebaker et~al.(2007)Pennebaker, Booth, and
  Francis}]{pennebakerlinguistic}
James~W Pennebaker, Roger~J Booth, and Martha~E Francis. 2007.
\newblock Linguistic inquiry and word count: {LIWC2007}.
\newblock \emph{Mahway: Lawrence Erlbaum Associates}.

\bibitem[{Pennington et~al.(2014)Pennington, Socher, and
  Manning}]{pennington2014glove}
Jeffrey Pennington, Richard Socher, and Christopher Manning. 2014.
\newblock Glove: Global vectors for word representation.
\newblock In \emph{Proceedings of EMNLP}.

\bibitem[{Qi et~al.(2018)Qi, Lin, Sun, Zhu, Xie, and Liu}]{qi2018cross}
Fanchao Qi, Yankai Lin, Maosong Sun, Hao Zhu, Ruobing Xie, and Zhiyuan Liu.
  2018.
\newblock Cross-lingual lexical sememe prediction.
\newblock In \emph{Proceedings of EMNLP}.

\bibitem[{Schuster and Paliwal(1997)}]{schuster1997bidirectional}
Mike Schuster and Kuldip~K Paliwal. 1997.
\newblock Bidirectional recurrent neural networks.
\newblock \emph{IEEE Transactions on Signal Processing}, 45(11):2673--2681.

\bibitem[{Sennrich et~al.(2016)Sennrich, Haddow, and
  Birch}]{sennrich2016neural}
Rico Sennrich, Barry Haddow, and Alexandra Birch. 2016.
\newblock Neural machine translation of rare words with subword units.
\newblock In \emph{Proceedings of ACL}.

\bibitem[{Thorat and Choudhari(2016)}]{thorat2016implementing}
Sushrut Thorat and Varad Choudhari. 2016.
\newblock Implementing a reverse dictionary, based on word definitions, using a
  node-graph architecture.
\newblock In \emph{Proceedings of COLING}.

\bibitem[{Tissier et~al.(2017)Tissier, Gravier, and
  Habrard}]{tissier2017dict2vec}
Julien Tissier, Christopher Gravier, and Amaury Habrard. 2017.
\newblock Dict2vec: Learning word embeddings using lexical dictionaries.
\newblock In \emph{Proceedings of EMNLP}.

\bibitem[{Wierzbicka(1996)}]{wierzbicka1996semantics}
Anna Wierzbicka. 1996.
\newblock \emph{Semantics: Primes and universals: Primes and universals}.
\newblock Oxford University Press, UK.

\bibitem[{Xie et~al.(2016)Xie, Liu, Jia, Luan, and Sun}]{xie2016representation}
Ruobing Xie, Zhiyuan Liu, Jia Jia, Huanbo Luan, and Maosong Sun. 2016.
\newblock Representation learning of knowledge graphs with entity descriptions.
\newblock In \emph{Proceedings of AAAI}.

\bibitem[{Xie et~al.(2017)Xie, Yuan, Liu, and Sun}]{xie2017lexical}
Ruobing Xie, Xingchi Yuan, Zhiyuan Liu, and Maosong Sun. 2017.
\newblock Lexical sememe prediction via word embeddings and matrix
  factorization.
\newblock In \emph{Proceedings of IJCAI}.

\bibitem[{Zeng et~al.(2018)Zeng, Yang, Tu, Liu, and Sun}]{zeng2018chinese}
Xiangkai Zeng, Cheng Yang, Cunchao Tu, Zhiyuan Liu, and Maosong Sun. 2018.
\newblock Chinese {LIWC} lexicon expansion via hierarchical classification of
  word embeddings with sememe attention.
\newblock In \emph{Proceedings of AAAI}.

\bibitem[{Zhang et~al.(2005)Zhang, Gong, and Wang}]{zhang2005chinese}
Yuntao Zhang, Ling Gong, and Yongcheng Wang. 2005.
\newblock Chinese word sense disambiguation using {HowNet}.
\newblock In \emph{Proceedings of International Conference on Natural
  Computation}.

\bibitem[{Zhong et~al.(2015)Zhong, Zhang, Wang, Wan, and
  Chen}]{zhong2015aligning}
Huaping Zhong, Jianwen Zhang, Zhen Wang, Hai Wan, and Zheng Chen. 2015.
\newblock Aligning knowledge and text embeddings by entity descriptions.
\newblock In \emph{Proceedings of EMNLP}.

\bibitem[{Zhu et~al.(2006)Zhu, Min, Zhou, Huang, and Wu}]{zhu2006semantic}
Yan-Lan Zhu, Jin Min, Ya-qian Zhou, Xuan-jing Huang, and Li-De Wu. 2006.
\newblock Semantic orientation computing based on {HowNet}.
\newblock \emph{Journal of Chinese Information Processing}, 20(1):14--20.

\end{thebibliography}
\bibliographystyle{acl_natbib}
\end{document}